\pdfoutput=1

\documentclass[11pt]{article}
\usepackage[preprint]{acl}

\usepackage{times}
\usepackage{latexsym}

\usepackage[T1]{fontenc}

\usepackage[utf8]{inputenc}

\usepackage{microtype}

\usepackage{inconsolata}

\usepackage{multirow}
\usepackage{graphicx}
\usepackage{booktabs}
\usepackage{algorithm}
\usepackage{algpseudocode}
\usepackage{amsmath}
\usepackage{amsfonts}
\usepackage{bm}
\usepackage[normalem]{ulem}
\usepackage{tabularx}
\usepackage{mathabx}  
\usepackage{colortbl}  
\usepackage{hyperref}
\usepackage{tikz}

\usepackage{diagbox}

\useunder{\uline}{\ul}{}

\def\eqref#1{equation~\ref{#1}}

\def\1{\bm{1}}

\def\vr{{\bm{r}}}

\def\vx{{\bm{x}}}

\DeclareMathAlphabet{\mathsfit}{\encodingdefault}{\sfdefault}{m}{sl}
\SetMathAlphabet{\mathsfit}{bold}{\encodingdefault}{\sfdefault}{bx}{n}

\newcommand{\ours}{SiGIR}

\newcommand{\oursb}{\underline{S}elf-Cr\underline{i}tique \underline{G}uided \underline{I}terative \underline{R}easoning}

\title{Self-Critique Guided Iterative Reasoning for
 Multi-hop Question Answering}

\author{
    Zheng Chu$^{1}$, 
    Huiming Fan$^{1}$, 
    Jingchang Chen$^{1}$,
    Qianyu Wang$^{1}$,
    Mingda Yang$^{1}$,\\
    \textbf{
    Jiafeng Liang$^{1}$,
    Zhongjie Wang$^{1}$,
    Hao Li$^{1}$,
    Guo Tang$^{1}$,
    Ming Liu$^{1,2}$\thanks{$\quad$ Corresponding Author.},
    Bing Qin$^{1,2}$}\\
    $^{1}$Harbin Institute of Technology, Harbin, China\\
    $^{2}$Peng Cheng Laboratory, Shenzhen, China\\
    \texttt{\{zchu,mliu\}@ir.hit.edu.cn}
}

\makeatletter
\def\therule{\makebox[\algorithmicindent][l]{\hspace*{.5em}\vrule height .75\baselineskip depth .25\baselineskip}}%

\newtoks\therules
\therules={}
\def\appendto#1#2{\expandafter#1\expandafter{\the#1#2}}
\def\gobblefirst#1{
  #1\expandafter\expandafter\expandafter{\expandafter\@gobble\the#1}}%
\def\LState{\State\unskip\the\therules}
\def\pushindent{\appendto\therules\therule}%
\def\popindent{\gobblefirst\therules}%
\def\printindent{\unskip\the\therules}%
\def\printandpush{\printindent\pushindent}%
\def\popandprint{\popindent\printindent}%

\algdef{SE}[WHILE]{While}{EndWhile}[1]
  {\printandpush\algorithmicwhile\ #1\ \algorithmicdo}
  {\popandprint\algorithmicend\ \algorithmicwhile}%
\algdef{SE}[FOR]{For}{EndFor}[1]
  {\printandpush\algorithmicfor\ #1\ \algorithmicdo}
  {\popandprint\algorithmicend\ \algorithmicfor}%
\algdef{S}[FOR]{ForAll}[1]
  {\printindent\algorithmicforall\ #1\ \algorithmicdo}%
\algdef{SE}[LOOP]{Loop}{EndLoop}
  {\printandpush\algorithmicloop}
  {\popandprint\algorithmicend\ \algorithmicloop}%
\algdef{SE}[REPEAT]{Repeat}{Until}
  {\printandpush\algorithmicrepeat}[1]
  {\popandprint\algorithmicuntil\ #1}%
\algdef{SE}[IF]{If}{EndIf}[1]
  {\printandpush\algorithmicif\ #1\ \algorithmicthen}
  {\popandprint\algorithmicend\ \algorithmicif}%
\algdef{C}[IF]{IF}{ElsIf}[1]
  {\popandprint\pushindent\algorithmicelse\ \algorithmicif\ #1\ \algorithmicthen}%
\algdef{Ce}[ELSE]{IF}{Else}{EndIf}
  {\popandprint\pushindent\algorithmicelse}%
\algdef{SE}[PROCEDURE]{Procedure}{EndProcedure}[2]
   {\printandpush\algorithmicprocedure\ \textproc{#1}\ifthenelse{\equal{#2}{}}{}{(#2)}}%
   {\popandprint\algorithmicend\ \algorithmicprocedure}%
\algdef{SE}[FUNCTION]{Function}{EndFunction}[2]
   {\printandpush\algorithmicfunction\ \textproc{#1}\ifthenelse{\equal{#2}{}}{}{(#2)}}%
   {\popandprint\algorithmicend\ \algorithmicfunction}%
\makeatother
\definecolor{bluegray}{HTML}{647D87}
\definecolor{Azure1}{HTML}{F0FFFF}
\definecolor{Burlywood4}{HTML}{8B7355}

\newcommand{\promptfigure}[2]{
\begin{tikzpicture}[
    every node/.style={outer sep=0},
    window/.style={rectangle, draw=black, rounded corners, thick, align=left, font=\small, inner xsep=0pt, inner ysep=6mm},
    titlestyle/.style={text=white},
]
\node[window] (window) at (0,-0.2cm) [minimum width=\linewidth, text width=\linewidth-6mm, anchor=north] {#2};
\draw[draw=black, fill=bluegray, thick] (-\linewidth/2,-0.35cm)
    [sharp corners] -- (\linewidth/2, -0.35cm)
    [rounded corners=5pt] -- (\linewidth/2, 0.35cm)
    [rounded corners=5pt] -- (-\linewidth/2, 0.35cm)
    [sharp corners] -- cycle;
\node[titlestyle] (title) at (0,-0.4mm) {#1};
\end{tikzpicture}%
}

\begin{document}
\maketitle

\begin{abstract}
Although large language models (LLMs) have demonstrated remarkable reasoning capabilities, they still face challenges in knowledge-intensive multi-hop reasoning.
Recent work explores iterative retrieval to address complex problems.
However, the lack of intermediate guidance often results in inaccurate retrieval and flawed intermediate reasoning, leading to incorrect reasoning.
To address these, we propose \oursb{} (\ours{}), which uses self-critique feedback to guide the iterative reasoning process.
Specifically, through end-to-end training, we enable the model to iteratively address complex problems via question decomposition. 
Additionally, the model is able to self-evaluate its intermediate reasoning steps.
During iterative reasoning, the model engages in branching exploration and employs self-evaluation to guide the selection of promising reasoning trajectories.
Extensive experiments on three multi-hop reasoning datasets demonstrate the effectiveness of our proposed method, surpassing the previous SOTA by 8.6\%.
Furthermore, our thorough analysis offers insights for future research.
Our code, data, and models are available at Github\footnote{\url{https://github.com/zchuz/SiGIR-MHQA}}.




\end{abstract}
\section{Introduction}
\label{sec:intro}
With the advancement of large-scale pre-training techniques, LLMs have achieved remarkable performance on reasoning tasks with chain-of-thought (CoT) prompting~\citep{fewshot-cot,survey-cot,gpt-4,deepseek-v2}.
Moreover, when augmented with \textit{Slow Thinking} mechanisms, several advanced models have demonstrated reasoning capabilities that either surpass or are comparable to those of human experts~\citep{deepseek-v3, o1}.

Despite their remarkable capabilities, LLMs are prone to generating factually inaccurate responses in knowledge-intensive tasks, primarily attributable to their limited access to domain-specific knowledge.
This limitation becomes particularly pronounced in multi-hop question answering (MHQA)~\citep{hotpotqa,2wikimhqa,musique}, which substantially compromises both the reasoning accuracy and factual reliability of the model's outputs.
Furthermore, the complexity of multi-hop questions presents additional challenges for global planning during the reasoning process



\begin{figure}[t]
    \centering
    \includegraphics[clip, width=\linewidth]{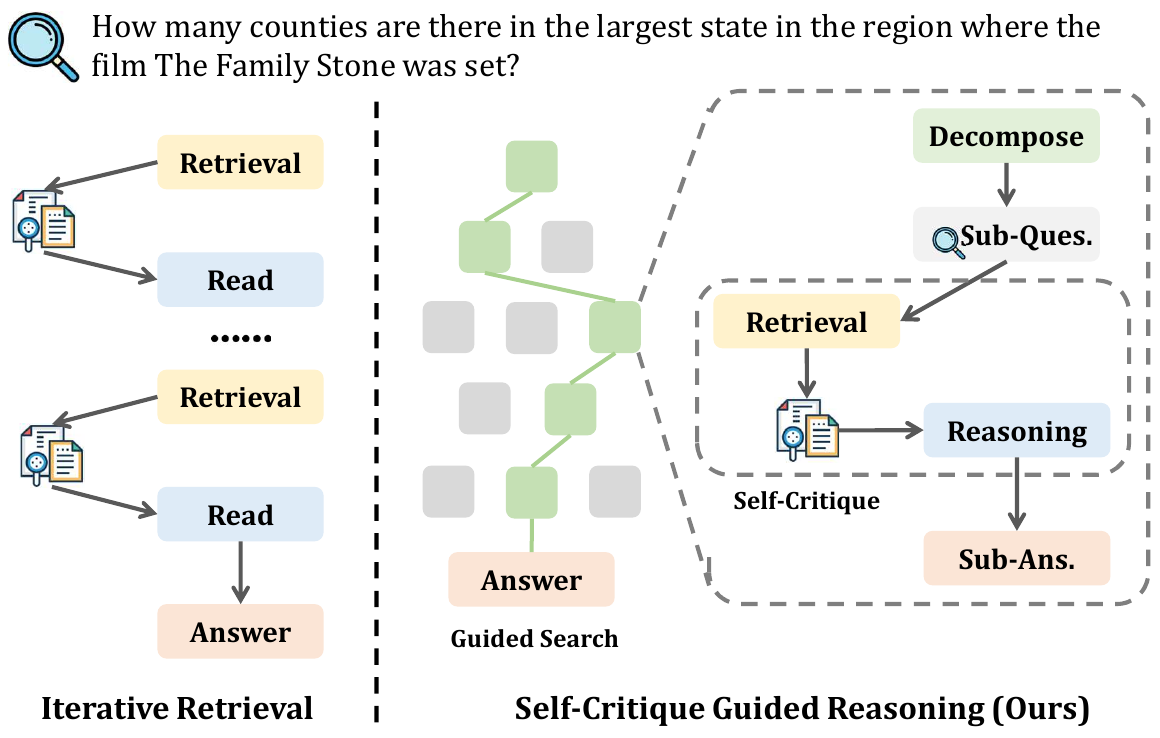}
    \caption{
    A comparison of iterative retrieval and self-critique guided reasoning (ours) in MHQA.
    Our method incorporates self-critique during iterative reasoning and conduct searches based on the received feedback.
    }
\label{fig:intro_figure}
\end{figure}

Retrieval-augmented generation is a promising strategy for mitigating the knowledge limitations of LLMs~\citep{rag-star}.
However, single-round retrieval is insufficient to encompass all relevant evidences to multi-hop questions~\citep{retrieval-then-read-1}.
To address this, iterative retrieval has been proposed~\citep{ircot,selfrag}.
By alternating between retrieval and generation, it can acquire more comprehensive relevant knowledge , improving retrieval performance.
Besides, some research focuses on the problems from the viewpoint of decomposition~\citep{ptot,beamaggr,semi-struct}.
It breaks down the complex multi-hop  question into simpler sub-questions, thereby reducing the difficulty of reasoning and planning, and enabling more accurate retrieval.


Although the aforementioned methods have partially addressed some aspects of challenges, they still suffer from certain limitations.
First, decomposing a complex question into sub-questions at the beginning of reasoning is highly  challenging,
and any mistakes can cause subsequent reasoning to deviate from the correct path.
Second, the iterative retrieval approach struggles to handle complex problem planning and articulate retrieval intents clearly, resulting in inaccurate retrieval and reasoning.
Finally, the absence of guidance for intermediate steps makes these methods susceptible to errors during reasoning, resulting in cascading errors.


To address these, our research focuses on following key aspects: 
(i) \textit{iterative question decomposition} facilitates complex question planning and accurate retrieval. 
(ii) \textit{intermediate feedback guided reasoning} expands the reasoning space and reduces cascading errors.
Based on this motivation, we introduce \oursb{} (\ours{}).
\ours{} iteratively solves complex questions through sub-question decomposition and selects promising reasoning paths guided by the self-critique feedback of intermediate reasoning steps.

Specifically, during \textit{\textbf{training phase}}, we synthesize iterative reasoning rationales with intermediate supervision from advanced LLMs. 
Thereafter, the model is equipped with the ability to perform question decomposition, trigger retrieval, knowledge reasoning, self-evaluation, and question reduction by means of end-to-end training.
During \textit{\textbf{inference phase}}, the model conducts iteration-level (step-level) beam search guided by self-feedback, as illustrated in Figure~\ref{fig:intro_figure}.
At each reasoning step, the model performs branch expansion for sub-question decomposition, retrieval, and reasoning. 
It then self-evaluates the current reasoning status and retain the top-k promising candidates based on self-critique rewards.
The model continues this iterative reasoning process until the final answer is derived.
Finally, the reasoning trajectory with the highest cumulative reward is selected as the final answer.


We conduct extensive experiments 
on three knowledge-intensive MHQA datasets: 
HotpotQA, 2WikiMQA, and MuSiQue~\citep{hotpotqa,2wikimhqa,musique}.
DeepSeek-V2.5,
Mistral,
LLaMA2,
and Qwen2.5 are used during experiments ~\citep{deepseek-v2,mistral,llama2,qwen25}.
Our method demonstrates significant advantages, with an average improvement of 8.6\% compared to the SOTA method.
Furthermore, the thorough analysis highlights the effectiveness of our method in retrieval, self-critique, search, and efficiency.

\section{Related Work}
\label{sec:related_work}

\subsection{Retrieval-Augmented Generation}
Retrieval-augmented generation (RAG) is an effective approach for mitigating hallucinations in LLMs~\citep{survey-rag,survey-hallucinnation}.
Retrieval-then-read, a widely-used approach in RAG, consists of query-based retrieval followed by conditional generation~\citep{retrieval-then-read-1,retrieval-then-read-2}.
However, its effectiveness is critically reliant on the quality of retrieval.
To address this, various methods have been proposed.
Query rewriting improves retrieval quality by reformulating retrieval queries~\citep{query-rewriting-1,query-rewriting-2,query-rewriting-3}.
Active retrieval dynamically adjusts retrieval timing and intent using adaptive mechanisms~\citep{active-retrieval-1,selfrag}.
Multi-round retrieval enables the acquisition of comprehensive information relevant to the problem~\citep{ircot,multi-round-retrieval-1}.

Our proposed method employs multi-round active retrieval to acquire more precise and comprehensive content for complex multi-hop questions.

\subsection{Multi-hop Question Answering}
MHQA requires gathering multiple pieces of evidence and systematic multi-step reasoning to arrive at the final answer.
Generally, single-round retrieval is inadequate  to acquire comprehensive information~\citep{retrieval-then-read-1,retrieval-then-read-2}.
Some methods adopt question decomposition to enhance accurate retrieval and reasoning~\citep{ptot,beamaggr,semi-struct}.
However, decomposing complex questions from the very beginning is particularly challenging.
To this end, iterative reasoning approach is introduced, which employs iterative multi-round reasoning to gradually tackle complex problems ~\citep{ircot,auto-rag,dr-distillation}.

In this work, we utilize iterative question decomposition and reduction to simplify the decomposition process and improve retrieval effectiveness.
\begin{figure*}[t]
    \centering
    \includegraphics[clip, width=\linewidth]{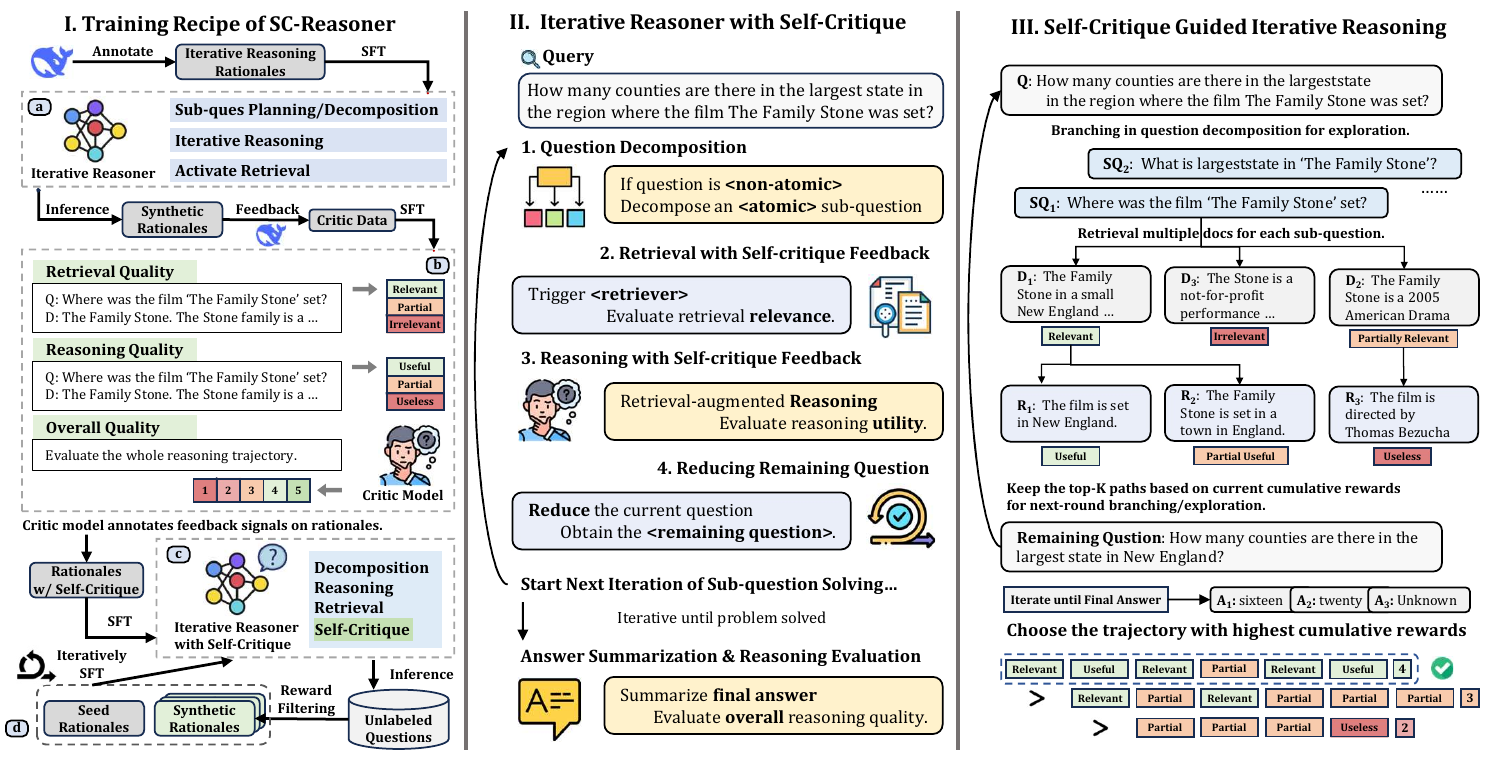}
    \caption{
    Overview of \ours{}.
    (I) Training recipe to get the iterative reasoner with self-critique, $R_{sc}$.
    (II) Reasoning characteristic of SC-Reasoner, which includes question decomposition, trigger retrieval, retrieval-augmented reasoning, self-evaluation, and question reduction.
    (III) SC-Reasoner engages in exploration during the iterative reasoning process and conducts reward-guided searches to select promising reasoning trajectories.
    }
\label{fig:method}
\end{figure*}

\subsection{Inference-Time Scaling}

Recent studies have found that increasing computational resources at inference can improve model performance, termed  inference-time scaling~\citep{inference-time-scaling-1,inference-time-scaling-2}.
\citet{self-consistency} employs a sampling ensemble technique to scale up, but the improvement is limited.
Other methods adopt search strategies, such as Monte Carlo Tree Search, which yield significant improvements, but involve higher computational overhead~\citep{mcts-1,mcts-2,mcts-3}.

In contrast, we introduce reward-guided search that circumvents the overhead that simulations entail.
It improves performance while ensuring reasonable costs, leading to efficient reasoning.




\section{Self-Critique Guided Iterative Reasoning for Multi-hop QA}
\label{sec:methods}

In this paper, we introduce \underline{S}elf-Cr\underline{i}tique \underline{G}uided \underline{I}terative \underline{R}easoning (SiGIR) framework for multi-hop question answering, as illustrated in Figure~\ref{fig:method}.
First, we synthesize iterative reasoning trajectories with intermediate supervision signals to train the model, which enables it to give self-critique of intermediate steps during reasoning.
During reasoning, the model iteratively explores and selects the optimal trajectory guided by self-critique feedback.

In this section, we will outline the data synthesis pipeline and training recipe in Section~\ref{method:data_synthesis_training},
introduce the inference pattern of self-critique iterative reasoner in Section~\ref{method:reasoner_with_self_critique},
and elaborate on the self-critique guided reasoning process in Section~\ref{method:inference_process}.

\subsection{Training Recipe}
\label{method:data_synthesis_training}

Our goal is to enable the model to perform iterative reasoning and to conduct self-critique of the intermediate reasoning steps.
To achieve this, we first  train an iterative reasoner $R$, followed by a critic model $C$.
Subsequently, we use $R$ and $C$ to generate rationales and train a model $R_{sc}$ capable of iterative reasoning and self-critique.
Figure~\ref{fig:method}(I) and Algorithm~\ref{alg:sc-reasoner-training} illustrate the training procedure.\footnote{Notations of training and inference are shown in Table~\ref{tab:notations}.}

\textbf{Iterative Reasoner.} We use advanced LLMs to synthesize iterative reasoning rationales.
Specifically, given a corpus $X$, where each instance $x_i$ includes a question $q_i$, an answer $a_i$, and corresponding documents $\{d_1,\cdots,d_n\}$.
For each instance, we use few-shot prompts with the question and supporting documents as input to the LLM, which yields question decompositions and sub-question reasoning.
Subsequently, we organize the sub-question decomposition, retrieval, and reasoning in an interleaved manner, and incorporate retrieval triggers before retrieved documents.
We train the smaller model with the parsed training instances to get the vanilla iterative reasoner $R$.
\footnote{Prompts for data synthesis can be found in Appendix~\ref{appendix:a3}.}

\textbf{Critic.} Due to the high cost of invoking LLMs for reward evaluation, we attempt to use a small model to clone the reward evaluation of LLMs.
To this end, we use iterative reasoning trajectories to construct the data for critic training.
Notably, to prevent the critic from overfitting to the generator's training data, we select a small amount of non-overlapping corpus with $X$ for critic data synthesis, denoted as $X_{critic}$.
Additionally, we employ $R$ instead of the LLM for reasoning synthesis, as $R$ tends to make mistakes during reasoning.
This helps ensure a balance between positive and negative samples.
Subsequently, we employ the LLM to evaluate rewards on retrieval, reasoning, and overall quality, obtaining training data for critic model.
We use it to train a separate  critic model $C$.

\textbf{Self-Critique Reasoner.} We aim to incorporate the critique signals into the reasoning process. 
Besides, recent studies have found that the generator and critic can enhance each other, and merging them into a single model can reduce inference cost.
To this end, we use the critic $C$ to annotate the sub-processes of the reasoning trajectories.
For retrieval evaluation, we take the query and retrieved documents as input, with retrieval relevance as outputs.
For sub-question reasoning, the inputs include sub-questions, documents, and reasoning.
For overall reward, we take the entire reasoning as input, and the output is the overall quality.
Subsequently, we add special tokens following the sub-processes, enabling the model to generate self-critique signals throughout the reasoning process.
We train the model with this corpus to obtain the reasoner with self-critique, denoted as SC-Reasoner, or $R_{sc}$.

SC-Reasoner is capable of reasoning and self-evaluation, facilitating data augmentation. 
Specifically, it can conduct reasoning on unlabeled questions and evaluate the quality of synthetic rationales, 
retaining high-quality rationales for subsequent training.
This self-improvement pipeline can improve the performance in data scarcity scenarios.
We use the standard causal language modeling objective for optimization during training.

\newcommand{\floatingtext}[1]{\kern-0.2em\makebox[0pt][l]{#1}}
\newcommand{\up}[1]{\floatingtext{\scriptsize($\uparrow$#1)}}

\begin{table*}[t]
\centering
\setlength\tabcolsep{4.5pt}
\resizebox{\textwidth}{!}{%
\begin{tabular}{lcccccccccccc}
\toprule
\multirow{2.5}{*}{\textbf{Methods}} & \multicolumn{5}{c}{\textbf{2WikiMQA}} & \multicolumn{3}{c}{\textbf{HotpotQA}} & \multicolumn{4}{c}{\textbf{MuSiQue}}  \\
\cmidrule(lr){2-6}  \cmidrule(lr){7-9} \cmidrule(lr){10-13}
 & \textbf{Overall} & \textit{Brg.} & \textit{Infer.} & \textit{Comp.} & \textit{B.C.} & \textbf{Overall} & \textit{Brg.} & \textit{Comp.} & \textbf{Overall} & \textit{2hop} & \textit{3hop} & \textit{4hop} \\
\midrule
\multicolumn{13}{l}{\textbf{\textit{Inference-based Methods}}} \\
Chain-of-thought$^*$ & 31.45 & 11.54 & 31.56 & 51.90 & 49.67 & 35.31 & 29.97 & 54.94 & 18.93 & 22.57 & 16.34 & 12.49  \\
Self-Ask$^*$ & 28.20 & 11.39 & 26.16 & 48.04 & 41.97 & 30.49 & 24.12 & 53.90  & 15.91 & 20.81 & 12.27 & 7.56 \\
RAG & 40.21 & 19.22 & 26.02 & 69.25 & 55.25 & 53.42 & 50.33 & 65.49 & 22.06 & 30.24 & 16.64 & 8.32  \\
IRCoT & 46.80 & 29.89 & 25.55 & 73.13 & 61.44 & 49.03  & 45.33 & 63.45 & 21.43  & 27.15 & {\ul 18.73} & 9.86 \\
ProbTree$^{\dagger}$ & 59.90 & - & - & - & - & 50.40 & - & - & 27.00 & - & - & - \\
BeamAggR$^{\dagger}$ & 63.20 & - & - & - & - & 55.20 & - & - & {\ul 32.30} & - & - & - \\
\midrule
\multicolumn{13}{l}{\textbf{\textit{Training-based Methods}}} \\
Self-RAG & 31.60 & 16.91 & 14.06 & 50.39 & 49.00 & 54.60  & 52.51 & 62.75 & 22.00  & 30.04 & 13.21 & {\ul 14.77} \\
RAG-Star$^{\dagger}$ & 46.80 & - & - & - & - & {\ul 60.00} & - & - & 30.70 & - & - & - \\
GenGround$^{\dagger}$ & 50.21 & - & - & - & - & 52.26 & - & - & 27.36 & - & - & - \\
Auto-RAG & 56.06 & 44.12 & 40.94 & 79.44 & 59.48 & 50.03 & 46.96 & 62.02 & 22.08 & 29.32 & 15.90  & 11.54 \\
DR-Distillation &  {\ul 70.51} & {\ul 46.13} & {\ul 56.34} & \textbf{96.90} & \textbf{96.00} & 58.06 & {\ul 54.08} & {\ul 73.62} & 22.74 & {\ul 35.08} & 12.06 & 6.57 \\
\midrule
\textbf{Ours} & \textbf{74.47} & \textbf{56.39} & \textbf{58.69} & {\ul 95.47} & {\ul 94.94} & \textbf{63.09} & \textbf{59.45} & \textbf{77.37} & \textbf{37.15} & \textbf{49.51} & \textbf{27.55} & \textbf{18.37} \\
~~- Qwen2.5 & 70.51 & 54.98 & 40.39 & 94.90 & 90.72 & 60.93 & 56.59 & 77.87 & 33.29 & 40.11 & 28.18 & 22.90 \\
~~- LLaMA2 & 65.82 & 50.93 & 43.36 & 90.44 & 79.25 & 52.92 & 47.92 & 72.46 & 28.56 & 39.00 & 20.60 & 12.98 \\
\bottomrule
\end{tabular}%
}
\caption{Experimental results on three multi-hop reasoning datasets: 2WikiMQA, HotpotQA, and MuSiQue.
Mistral-7B is used as the default backbone.
Best and second results are highlighted by \textbf{bold} and {\ul underline}. 
The evaluation metric is F1.
$*$ indicates closed-book reasoning, 
and $\dagger$ denotes results are derived from the original paper.
}
\label{tab:main_results}
\end{table*}

\subsection{Iterative Reasoner with Self-Critique}
\label{method:reasoner_with_self_critique}

This section introduces the reasoning patterns of SC-Reasoner.
Overall, SC-Reasoner is capable of question decomposition and planning, triggering retrieval, reasoning, and self-evaluating the reasoning process.
As illustrated in Figure~\ref{fig:method}(II), given a question, the model first determines whether the question needs decomposition.
If  needed, the model decomposes one atomic sub-question at a time.
Then, the model triggers external retrieval and conducts knowledge reasoning, along with self-evaluating the quality of both retrieval and reasoning.

Whenever the model resolves a sub-question, it reduces the original question to identify the remaining unresolved part, thereby simplifying the overall question.
The model iteratively repeats this process until the reduced atomic question is resolved, leading to the final answer.
At this point, it conducts self-evaluation of the entire reasoning trajectory.

\subsection{Self-Critique Guided Iterative Reasoning}
\label{method:inference_process}

After training, SC-Reasoner can conduct iterative reasoning and self-critique. 
Accordingly, we introduce self-critic guided iterative reasoning, 
an iteration-level beam search algorithm.
This algorithm searches within the iterative reasoning process, employing self-rewards to guide the selection of promising reasoning candidates, as illustrated in Figure~\ref{fig:method}(III) and Algorithm~\ref{alg:sc-reasoner-inference}.
Generally, in each iteration, the model first expands the exploration space by branching based on the current reasoning prefix, and then selects top-k candidate trajectories guided by self-rewards.
At the end of the reasoning, we use self-rewards to choose the final answer.

\paragraph{Branch Exploration}
Each iteration consists of the following steps:
question decomposition, retrieval, reasoning, remaining question reduction, and self-critique.
At timestamp $t$, the model branches out and expands downward through temperature sampling in question decomposition, retrieval, and reasoning, as shown in equations below.
\begin{align}
    q_{t+1} &\sim p(y \mid q_{t}^{(r)}, x_{<t}) \\
    s_{t+1} &\sim p(y \mid q_{t+1}, d_i, x_{<t})
\end{align}
where $\sim$ is sampling, $q_t^{(r)}$ is remaining question,  $q$ is sub-question, $s$ is sub-question reasoning, $p$ is probabilistic LM, and $d$ is retrieved documents.

During the aforementioned process, the model also conducts self-evaluation and assigns rewards, as shown in Equ. (2,3).
At the end of an iteration, the model will reduce the question for the next iteration based on current reasoning status $x_{t+1}$.
\begin{align}
    r_{t+1}^{\mathrm{retr}} &= p(y \mid q_{t+1}, d_i) \\
    r_{t+1}^{\mathrm{reas}} &= p(y \mid q_{t+1}, d_i, s_{t+1}) \\
    x_{t+1} &= \oplus[x_t, q_{t+1}, d, r_{t+1}^{\mathrm{retr}}, s_{t+1}, r_{t+1}^{\mathrm{reas}}] \\
    q_{t+1}^{(r)} &= p(y \mid x_{t+1})
\end{align}

\paragraph{Candidate Selection}
Through branch exploration, the model's reasoning investigates a wide space.
Subsequently, we aim to select the most promising candidates among them based on rewards.
Selected candidates will serve as the starting point for the  next iteration of reasoning.
Specifically, we use the cumulative process reward as the criterion for candidate selection.
The cumulative reward consists of reward from previous iteration and current iteration.
We retain the candidates with top-k largest rewards, as shown in Equ. (7, 8).
Once the reasoning is finished, we need to select one candidate as the final answer, 
using either cumulative reward of entire trajectory or outcome reward.
\begin{align}
    r^{c}_{t+1} &= 
    \underbrace{r^{c}_{t}}_{\text{previous reward}}
     +
    \underbrace{r_{t+1}^{\mathrm{retr}} + r_{t+1}^{\mathrm{reas}}}_{\text{current reward}} \\
    \vx_{t+1}^{*} &= \text{top-k}(
    \vx_{t+1},  \vr_{t+1}^{c}
    ) \\
    r^{c} &= 
    \sum_{i=1}^{n}
    (r_i^{\mathrm{retr}} + r_i^{\mathrm{reas}}) \\
    r^{o} &= 
    p(y \mid q^{(r)}, x_{<n}, a) 
\end{align}
where $r^c$ and $r^o$ denote cumulative process reward and outcome reward, respectively.
\section{Experimental Setup}
\label{sec:setup}

\subsection{Benchmarks}
We evaluate \ours{} on three knowledge-intensive multi-hop reasoning datasets: HotpotQA~\citep{hotpotqa}, 2WikiMQA~\citep{2wikimhqa}, and MuSiQue~\citep{musique}.
Following previous work~\citep{ircot,ptot}, we use the development set as the test data.
The experiments are conducted in the open-domain setting.
We use token-level F1 as the evaluation metric.


\subsection{Baselines}
We select extensive baselines for comparison, including the following methods: 
Fewshot CoT~\citep{fewshot-cot}, 
Self-Ask~\citep{selfask}, 
IRCoT~\citep{ircot}, 
ProbTree~\citep{ptot}, 
BeamAggR~\citep{beamaggr},
RAG~\citep{selfrag}, 
RAG-Star~\citep{rag-star}, GenGround~\citep{generate-then-ground}, 
Auto-RAG~\citep{auto-rag}, 
and DR-Distillation~\citep{dr-distillation}.
\footnote{Details about benchmarks, retrieval, baselines, implementation, and task definition can be found in Appendix~\ref{appendix:a1}.}

\subsection{Implementation Details}
We use DeepSeek-v2.5~\citep{deepseek-v2} as the advanced LLM in our experiments.
By default, we use BM25~\citep{bm25} for sparse retrieval. 
In analysis, we additionally use Contriever~\citep{contriever} for dense retrieval.
During training, we use LoRA~\citep{lora} for parameter-efficient fine-tuning, with Huggingface Accelerate for mixed-precision training.
The entire training process takes approximately 3 hours on 4 $\times$ A100-SXM4-80GB GPUs.

\section{Experimental Results}
\label{sec:results}

We conduct extensive experiments on three multi-hop reasoning datasets, as shown in Table~\ref{tab:main_results}.
We notice that although CoT is considered proficient in multi-step reasoning, but its performance is poor due to a lack of question-relevant knowledge within the LLM.
When single-round retrieval is employed, an average performance improvement of 35.0\% is observed.
Iterative multi-round retrieval results in improved performance, achieving up to a 48.8\% increase on 2WikiMQA.
Additionally, methods based on question decomposition further improve reasoning performance.
These  results suggest that precise retrieval and reasoning planning are key factors affecting multi-hop reasoning.

Some methods equip the model with iterative reasoning capabilities through fine-tuning, introducing strategies such as active retrieval, question decomposition, and search during the iterative reasoning process.
These methods can better address complex problems and achieve more accurate retrieval, further enhancing the performance.

\begin{table}[t]
\centering
\setlength\tabcolsep{4pt}
\resizebox{\linewidth}{!}{%
\begin{tabular}{lccc}
\toprule
\textbf{Setttings} & \textbf{2WikiMQA} & \textbf{HotpotQA} & \textbf{MuSiQue} \\
\midrule
\textbf{SGIR} & \textbf{74.47}  & \textbf{63.09} & \textbf{37.15} \\
\midrule

\multicolumn{4}{l}{\textbf{\textit{Fine-grained Reward}}} \\
(a) w/o. guided search & 72.37 & 60.86 & 35.47 \\
(b) w/o. reward & 67.23 & 54.94 & 29.45 \\
(c) w/o. self-critique & 66.96 & 56.86 & 28.62 \\
\midrule
\multicolumn{4}{l}{\textbf{\textit{Coarse-grained Reward}}} \\
(d) w. coarse reward & 73.40 & 59.85 & 34.32 \\
(e) w/o. guided search & 72.32 & 60.89 & 35.40 \\
\bottomrule
\end{tabular}%
}
\caption{
Ablation experiments on rewards and search algorithm in fine-grained and coarse-grained scenarios.
The experiments are conducted with Mistral-7B.
}
\label{tab:analysis_ablation}
\end{table}

As shown in Table~\ref{tab:main_results}, our method demonstrates remarkable superiority over baselines.
It surpasses the previous state-of-the-art methods on all three datasets, achieving improvement of 2WikiMQA (+3.96), HotpotQA (+3.09), and MuSiQue (+4.85).
Furthermore, our method excels particularly on the challenging MuSiQue dataset, which involves 3 to 4 hop hard questions.
It achieves a significant improvement of 15.0\% on the MuSiQue dataset compared to previous methods, with enhancements of 47.1\% and 24.37\% on 3/4-hop challenging questions, respectively.
Additionally, we also conduct experiments on the Qwen2.5 and LLaMA2 models, achieving competitive performance that demonstrates the generalizability of our method.

We identify three key factors that contribute to our improvement:
(a) Iterative multi-round reasoning and sub-question planning reduce the complexity of questions while improving retrieval accuracy.
(b) Self-critique provides valuable feedback signals for intermediate steps during the reasoning process.
(c) Reward-guided search selects superior reasoning trajectories among the exploration space.

\section{Analysis}
\label{sec:analysis}

\subsection{Ablation Study}
\label{analysis:ablation_analysis}

We conduct ablation experiments to investigate the effects of different modules, as shown in Table~\ref{tab:analysis_ablation}.
Our method employs fine-grained rewards by default.
First, removing reward-guided search (a) and substituting it with greedy inference leads to considerable performance declines on all three datasets, highlighting the importance of search algorithms.
Next, we remove the reward signal from greedy inference (b), which means the model's inference process would not receive any guidance.
This change results in a substantial performance decrease of 10\%, demonstrating the essential role of reward signals in the inference process.
Additionally, the exclusion of self-critique data from the training data produces a minor performance decline.

To further investigate the role of reward signals, particularly the impact of reward granularity on model performance, we replace the rewards with coarse-grained binary rewards, as shown in Table~\ref{tab:analysis_ablation}.
In reward-guided greedy inference (e), fine-grained and coarse-grained rewards have similar effects.
However, when a broader search is required (d), coarse-grained rewards exhibit certain drawbacks, resulting in a 2.7\%  performance decrease compared to fine-grained rewards. 
This indicates that fine-grained rewards are more effective in search-form reasoning, particularly when there are a substantial number of intermediate reasoning steps.

\begin{figure}[t]
    \centering
    \includegraphics[clip, width=\linewidth]{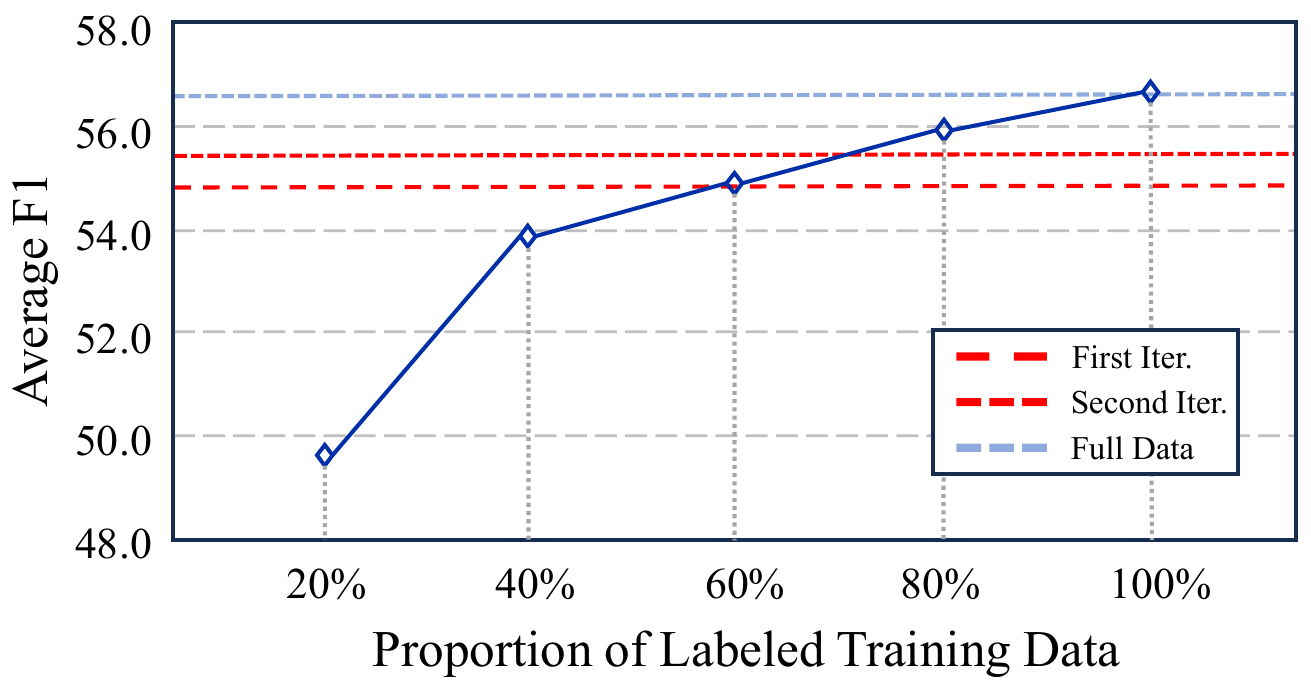}
    \caption{
    The impact of increasing training data and self-improvement on performance.
    As data scales, the performance consistently improves.
    Starting with 40\% of the data, two self-improvement iterations boost performance to nearly 80\%, with only 2\% gap with full sft.  
    }
\label{fig:self-improvement}
\end{figure}
\begin{table}[t]
\centering
\setlength\tabcolsep{4pt}
\resizebox{\linewidth}{!}{%
\begin{tabular}{lcccc}
\toprule
\textbf{Stage } & \textbf{\#Labeled} & \textbf{\#Unlabeled} & \textbf{\#Total} & \textbf{Quantile}\\
\midrule
Training Set & 49000 & - & - & -\\
Full SFT & 38625 & 0 & 38625 & 100\% \\
SP Start Point & 15000 & 0 & 15000 & 95.3\% \\
First Iteration & 15000 & 21669 & 36669 & 97.1\% \\
Second Iteration & 15000 & 29659 & 44659 & 97.8\% \\
\bottomrule
\end{tabular}%
}
\caption{
Comparison of the amount of data used for training and performance quantiles.
\#Labeled refers to data annotated by the teacher model with reasoning trajectories, 
and \#Unlabeled refers to data synthesized by the smaller model itself.
Quantile indicates the percentage of performance compared to full SFT.
}
\label{tab:analysis_seed_data}
\end{table}

\subsection{Training Scaling and Self-Improvement}
\label{analysis:self_improvement}

We investigate the performance under different amounts of training data and the effect of self-improvement data synthesis, as shown in Figure~\ref{fig:self-improvement}.
Initially, as the amount of labeled data increases, the performance improves consistently.
The most significant improvement occurs when 40\% of the training data is utilized; beyond this threshold, the rate of performance enhancement decreases.

Starting with the model trained on 40\% of the data, we perform iterative self-improvement.
As shown in Figure~\ref{fig:self-improvement} and Table~\ref{tab:analysis_seed_data}, in the first iteration of self-improvement, the model achieves performance corresponding to the 60\% training data quantile.
In the second iteration, the model's performance approaches the 80\% quantile, with only a 2.2\% gap from the performance achieved with the full labeled data fine-tuning.
This suggests that our approach can effectively synthesize data through self-critique, yielding competitive performance in situations where labeled data is scarce.

\begin{table}[t]
\centering
\setlength\tabcolsep{4pt}
\resizebox{\linewidth}{!}{%
\begin{tabular}{lcccc}
\toprule
\textbf{Strategy} & \textbf{2WikiMQA} & \textbf{HotpotQA} & \textbf{MuSiQue} & \textbf{Average}\\
\midrule
\multicolumn{5}{l}{\textbf{\textit{Search Strategy}}} \\
Guided Search & 74.47 & 63.09 & 37.15 & 58.23 \\
~~(a) one candidate & 72.97 & 61.27 & 36.21 & 56.81 \\
~~(b) no branching & 68.87 & 52.88 & 28.34 & 50.03 \\
Greedy Inference & 72.37 & 60.86 & 35.47 & 56.23 \\
\midrule
\multicolumn{5}{l}{\textbf{\textit{Selection Strategy}}} \\
(c) cumulative reward & 74.47 & 63.09 & 37.15 & 58.23 \\
(d) outcome reward & 71.79 & 60.96 & 35.23 & 55.99 \\
(e) random select & 71.17 & 60.10 & 35.17 & 55.81 \\
\bottomrule
\end{tabular}%
}
\caption{
Analysis on various search and voting strategies on our method.
Retaining more candidate reasoning trajectories and exploring different sub-questions help improve performance.
Utilizing cumulative process rewards to select the final reasoning trajectory is effective.
}
\label{tab:analysis_search}
\end{table}

\subsection{Search and Trajectory Selection}
\label{analysis:search_selection_strategy}

We conduct experiments to investigate the factors influencing the guided reasoning process, including search strategies and reasoning selection, as shown in Table~\ref{tab:analysis_search}.
As indicated in the preceding discussion, reward-based search is more effective than greedy inference.
During the search process, we retain only one candidate trajectory in each iteration (a), which results in a performance decrease of 2.4\%.
This indicates that maintaining more candidate sequences preserves a greater number of potentially correct reasoning trajectories.
Next, we set the question decomposition exploration width to 1 (b), which means only retrieval will be expanded during reasoning. 
This leads to a substantial drop of 13.57\%,
indicating the significance of exploring various sub-questions in the reasoning process.

At the end of the search, we need to select one trajectory as the final answer, and we investigate various answer selection strategies.
Using cumulative process rewards for trajectory selection yields the best performance, followed by using outcome rewards.
Both strategies outperform random selection.
This suggests that in complex reasoning processes, intermediate rewards can serve as a criterion for assessing the reasoning quality, and that process rewards 
are more effective at distinguishing between different levels of reasoning quality.

\begin{table}[t]
\centering
\setlength\tabcolsep{4pt}
\resizebox{\linewidth}{!}{%
\begin{tabular}{lcccc}
\toprule
\textbf{Reward / Task} & \textbf{Retrieval} & \textbf{Reasoning} & \textbf{Overall} & \textbf{Average} \\
\midrule
Direct Reward & 93.2 & \textbf{95.8} & 81.2 & 90.1 \\
Post Explanation & 95.4 & 95.4 & 80.6 & 90.5 \\
Chain-of-thought & \textbf{97.4} & 88.6 & \textbf{87.4} & \textbf{91.1} \\
~~- Dedicated RM & 95.8 & 93.6 & 78.8 & 89.4\\
\bottomrule
\end{tabular}%
}
\caption{
The performance of different reward forms on reward modeling tasks.
Generative rewards are more effective than discriminative rewards, and mixed reward modeling has advantages over dedicated reward models in complex outcome reward modeling task.
}
\label{tab:analysis_reward}
\end{table}

\begin{table}[t]
\centering
\setlength\tabcolsep{4pt}
\resizebox{\linewidth}{!}{%
\begin{tabular}{lccccc}
\toprule
\textbf{Strategy} & \textbf{2WikiMQA} & \textbf{HotpotQA} & \textbf{MuSiQue} & \textbf{Average} &  \textbf{Throughput} \\
\midrule
GenCritic & \textbf{72.37} & \textbf{60.86} & 35.47 & \textbf{56.23} & \textbf{1.92} QPS  \\
Seperated & 72.23 & 58.45 & \textbf{36.46} & 55.71 & 0.43 QPS  \\
\bottomrule
\end{tabular}%
}
\caption{
Comparison of performance and efficiency between generative and separated critic model.
The generative critic model brings slightly better performance, but has significant higher inference throughput.
}
\label{tab:analysis_seperate_critic}
\end{table}


\subsection{Reward Modeling }
\label{analysis:reward_modeling}
The forms of reward (critic) models are diverse, such as generative and discriminative forms.
We examine the performance of various types of critic models on our reward modeling task, as shown in Table~\ref{tab:analysis_reward}.
Overall, generative reward modeling outperforms discriminative modeling. 
Moreover, among generative forms, chain-of-thought (CoT) is preferred over post-hoc explanations.
Notably, CoT excels at modeling complex outcome rewards because it systematically decomposes the entire reasoning trajectory for step-by-step analysis.

Additionally, we investigate dedicated RM (each model assigned to a specific task) in comparison to mixed RM (one model responsible for all tasks).
The dedicated RM exhibits performance variations of -1.6\% / +5.6\% for retrieval and reasoning tasks, respectively. 
However, there is a significant performance decrease of 9.8\% in modeling outcome rewards. 
Since outcome rewards assess the entire reasoning process by integrating various reward modeling tasks, this suggests that multi-task reward modeling of sub-tasks enhances the model's ability to evaluate complex outcome rewards.


Furthermore, we highlight that generative critic models offer greater benefits than separated generators and critics.
As shown in Table~\ref{tab:analysis_seperate_critic}, the GenCritic outperforms the separated models in two of the three datasets, achieving an average performance improvement of 0.9\%.
Without sacrificing reasoning performance, the GenCritic achieves a throughput that is 4.46 times higher, demonstrating significantly better reasoning efficiency\footnote{Please refer to Appendix~\ref{a2:reward-modeling} for details of analysis.}.

\begin{table}[t]
\centering
\setlength\tabcolsep{4pt}
\resizebox{\linewidth}{!}{%
\begin{tabular}{lcccc}
\toprule
\textbf{Retriever} & \textbf{2WikiMQA} & \textbf{HotpotQA} & \textbf{MuSiQue} & \textbf{Average}\\
\midrule
\multicolumn{5}{l}{\textbf{\textit{Reward-guided Greedy Inference}}} \\
Sparse & 72.37 & 60.86 & 35.47 & 56.23 \\
Dense & 75.44 & 59.53 & 38.48 & 57.81\\
Hybrid & \textbf{76.04} & \textbf{62.89} & \textbf{39.38} & \textbf{59.43} \\
\midrule 
\multicolumn{5}{l}{\textbf{\textit{Reward-guided Search}}} \\
Sparse & 74.47 & 63.09 & 37.15 & 58.23 \\
Dense & 75.72 & 61.72 & \textbf{40.36} & \textbf{59.26} \\
Hybrid & \textbf{76.05} & \textbf{63.47} & 37.05 & 58.85 \\

\bottomrule
\end{tabular}%
}
\caption{
The impact of retrieval systems on reasoning.
Accurate retrieval leads to better reasoning performance.
Our method naturally supports hybrid retrieval (sparse + dense) retrieval, which facilitates better reasoning.
}
\label{tab:analysis_retrieval}
\end{table}

\begin{table}[t]
\centering
\setlength\tabcolsep{4pt}
\resizebox{\linewidth}{!}{%
\begin{tabular}{lcccc}
\toprule
\textbf{Retriever} & \cellcolor[HTML]{FFFDFA}{\color[HTML]{333333} \textbf{2WikiMQA}} & \cellcolor[HTML]{FFFDFA}{\color[HTML]{333333} \textbf{HotpotQA}} & \cellcolor[HTML]{FFFDFA}{\color[HTML]{333333} \textbf{MuSiQue}} & \cellcolor[HTML]{FFFDFA}{\color[HTML]{333333} \textbf{Average}} \\
\midrule

{\color[HTML]{333333} BM25} & {\color[HTML]{333333} 72.37} & {\color[HTML]{333333} 60.86} & {\color[HTML]{333333} 35.47} & {\color[HTML]{333333} 56.23} \\

{\color[HTML]{333333} Contriever} & {\color[HTML]{333333} 75.44} & {\color[HTML]{333333} 59.53} & {\color[HTML]{333333} 38.48} & {\color[HTML]{333333} 57.81} \\

{\color[HTML]{333333} + Hybrid} & {\color[HTML]{333333} 76.04} & {\color[HTML]{333333} 62.89} & {\color[HTML]{333333} 39.38} & {\color[HTML]{333333} 59.43} \\
\midrule

{\color[HTML]{333333} BGE-large} & {\color[HTML]{333333} 74.36} & {\color[HTML]{333333} 64.27} & {\color[HTML]{333333} 40.58} & {\color[HTML]{333333} 59.74} \\

{\color[HTML]{333333} + Hybrid} & {\color[HTML]{333333} 75.85} & {\color[HTML]{333333} 64.25} & {\color[HTML]{333333} 41.03} & {\color[HTML]{333333} 60.38} \\

\midrule
{\color[HTML]{333333} ColBERT-v2} & {\color[HTML]{333333} 75.37} & {\color[HTML]{333333} 65.44} & {\color[HTML]{333333} 42.24} & {\color[HTML]{333333} 61.02} \\

{\color[HTML]{333333} + Hybrid} & {\color[HTML]{333333} 75.37} & {\color[HTML]{333333} 65.69} & {\color[HTML]{333333} 43.05} & {\color[HTML]{333333} 61.37} \\
\bottomrule
\end{tabular}
}
\caption{
Experiments on different dense retrievers and their hybrid retrieval variants.
Hybrid retrieval strategies improve performance across various dense retrievers.
}
\label{tab:analysis_retrieval_additional}

\end{table}

\subsection{Retrieval System and Hybrid Retrieval}
\label{analysis:retrieval_hybrid}
Retrieval plays a critical role in knowledge reasoning.
To investigate the effect of the retrieval system, we employ various retrieval systems in the experiments, as illustrated in Table~\ref{tab:analysis_retrieval}.
Dense retrieval has better performance than sparse retrieval. When employing dense retrieval, a notable improvement of 2.8\% is observed, suggesting that retrieval quality impacts reasoning performance.
Furthermore, since our method treats each retrieved document independently, it naturally facilitates the integration of different retrieval systems.
The hybrid retrieval system
achieves a significant improvement of 5.7\%, demonstrating the superiority of our method.
However, in a guided search, excellent reasoning performance can be achieved using only sparse retrieval.
This suggests that when retrieval quality is poor, the reward-guided search can help the model in exploring and filtering irrelevant retrieved documents, thereby improving reasoning performance.

In addition, we also apply hybrid retrieval on different dense retrievers (BGE-large\footnote{\url{http://huggingface.co/BAAI/bge-large-en-v1.5}} and ColBERT-v2\footnote{\url{http://huggingface.co/colbert-ir/colbertv2.0}}), as shown in Table~\ref{tab:analysis_retrieval_additional}.
The hybrid retrieval strategy consistently achieves significant performance improvements across different dense retrievers, demonstrating its effectiveness.


\begin{figure}[t]
    \centering
    \includegraphics[clip, width=\linewidth]{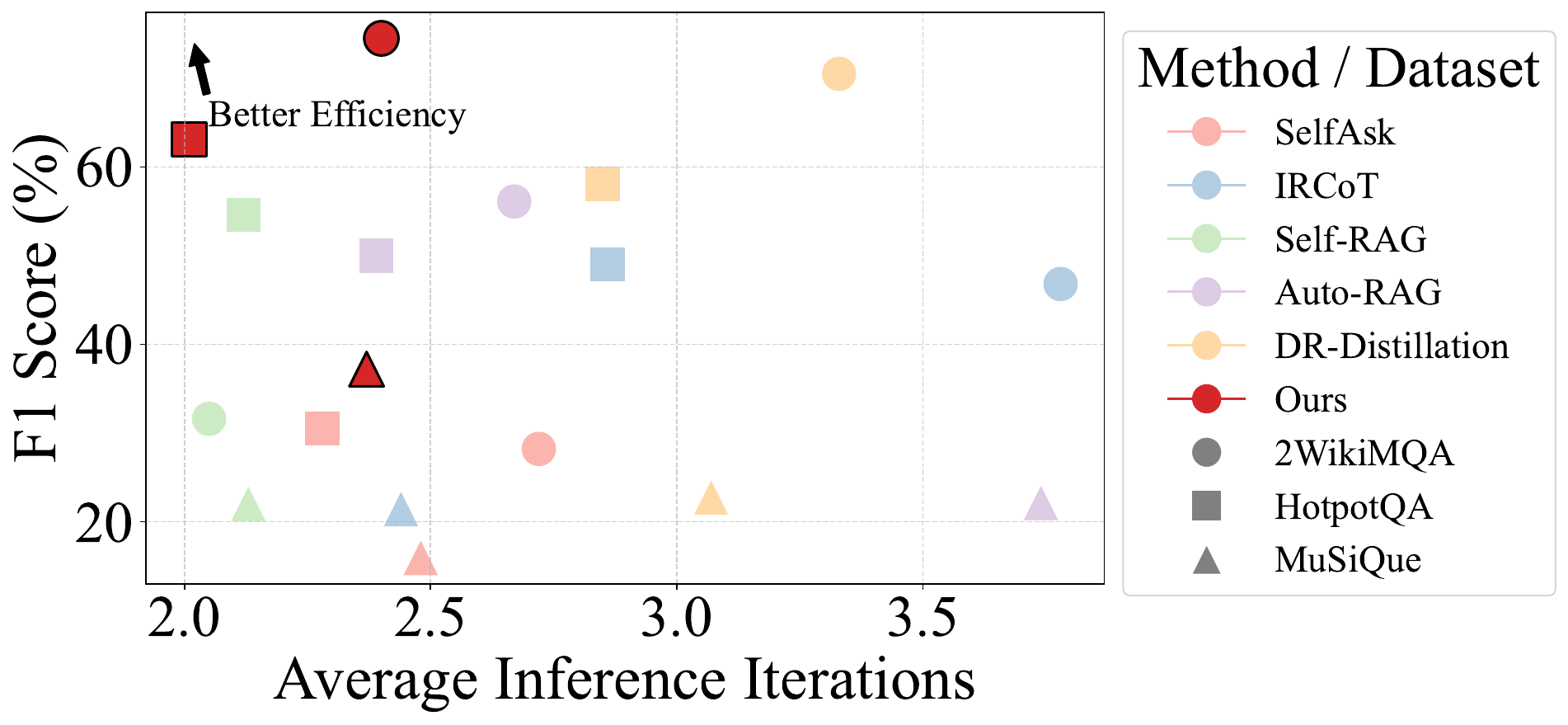}
    \caption{
    Relationship between performance and the average number of iterations to solve a multi-hop question.
    Upper-left represents efficient reasoning.
    Our method requires fewer iterations to solve one question with higher F1, demonstrating higher reasoning efficiency.
    }
\label{fig:analysis_pareto}
\end{figure}

\subsection{Iterative Reasoning Efficiency}
\label{analysis:efficiency}
We assess reasoning efficiency by measuring the iterations needed to solve a multi-hop problem alongside its reasoning performance, as illustrated in Figure~\ref{fig:analysis_pareto}.
Among the mentioned methods, Auto-RAG~/~DR-Distillation shows commendable reasoning capabilities, but it demands more iterations of reasoning, which incurs substantial costs.
Conversely, our method not only surpasses their performance but also reduces the average number of reasoning iterations by 26.7\%, demonstrating superior reasoning efficiency.
Additionally, we also measure the average inference latency of single samples and token consumption in Appendix~\ref{a2:iterative-reasoning-efficiency}.


\section{Conclusion}
\label{sec:conclusion}

In this work, we introduce \oursb{} for knowledge-intensive multi-hop question answering.
Initially, we enable the model to conduct iterative reasoning through iterative question decomposition and to self-evaluate the intermediate reasoning steps, through a novel data synthesis and training pipeline.
During inference, the model performs self-critique guided search:
it first explores multiple reasoning paths via branching exploration
and then utilizes self-critique feedback to refine the search process and identify the most promising reasoning trajectory.
Extensive experiments and thorough analysis demonstrate the effectiveness of our proposed method.

\section*{Limitations}
Although we have conducted comprehensive experiments to demonstrate the effectiveness of our proposed methods, and provided detailed analysis on the components of our system, 
this paper still has some limitations.

For sub-question exploration, we employ a sampling-based approach rather than training a dedicated question decomposition module.
Although this design improves the simplicity and efficiency of the overall system, it can limit the exploratory breadth of the model with respect to sub-questions.
We plan to address this in future research, such as incorporating a query rewriting module.

Our method adopts a route of synthetic data followed by supervised fine-tuning (SFT), without incorporating reinforcement learning in the model's sub-question reasoning process.
In the future, we consider using strategies such as Monte Carlo tree search to generate richer reasoning trajectories and introducing reinforcement learning to optimize these trajectories, thereby further enhancing the model's reasoning performance.
\section*{Acknowledgements}
The research in this article is supported by the National Key Research and Development Project (2022YFF0903301),  the National Science Foundation of China (U22B2059, 62276083). We also appreciate the support from China Mobile Group Heilongjiang Co., Ltd. @ on our research, the research is jointly completed by both parties. 
Ming Liu is the corresponding author.

\bibliography{custom}

\appendix
\section{Supplemental Materials on Setup}
\label{appendix:a1}

\subsection{Datasets}
\label{a1:datasets}
\paragraph{HotpotQA}~\citep{hotpotqa}\footnote{\url{https://hotpotqa.github.io/}} is a multi-hop reasoning dataset that includes two reasoning types: bridge and comparison.
Bridge: sub-questions are linked by bridging entities.
Comparison: compare two statements, including yes/no and partial order.

\paragraph{2WikiMQA}~\citep{2wikimhqa}\footnote{\url{https://github.com/Alab-NII/2wikimultihop}} is a multi-hop reasoning dataset that contains four reasoning types.
The bridge (compositional) and comparison are the same as those in HotpotQA.
Besides, it includes two additional reasoning types: bridge-comparison and inference.
Bridge-comparison: compare two bridge sub-questions.
Inference: sub-questions are connected through reasoning relationships rather than bridging entities. 
For example, the predicate grandfather is equal to a father's father.

\paragraph{MuSiQue}~\citep{musique}\footnote{\url{https://github.com/StonyBrookNLP/musique}} is a multi-hop reasoning dataset including 2-4 hops questions.
All questions in MuSiQue are compositional, with the number of hops extended to a maximum of 4 hops.
Besides, it includes multiple compositional questions, where a question contains more than one bridge entity.
Due to the complexity of the questions, MuSiQue has the highest level of difficulty.

\subsection{Models}
\label{a1:models}
We use DeepSeek-V2.5~\citep{deepseek-v2}\footnote{\url{https://huggingface.co/deepseek-ai/DeepSeek-V2.5}} as the advanced LLM to synthesis reasoning and critic rationales.
In the experiments, we use Mistral-7B~\citep{mistral}\footnote{\url{https://huggingface.co/mistral-community/Mistral-7B-v0.2}} as the student backbone model by default.
Besides, we also conduct experiments on LLaMA2-7B~\citep{llama2}\footnote{\url{https://huggingface.co/meta-llama/Llama-2-7b}} and Qwen2.5-7B~\citep{qwen25}\footnote{\url{https://huggingface.co/Qwen/Qwen2.5-7B}} to demonstrate the generalization of our method.

\subsection{Training Corpus}
\label{a1:training-corpus}
We randomly sample subsets of the training data from 2WikiMQA, HotpotQA, and MuSiQue to construct the training corpus.
Specially, we sample 20,000, 10,000, 19,000 instances from the respective datasets to construct the initial generator training corpus $X$, while an additional 3,000, 2,000, and 1,000 non-overlapping instances are sampled as the critic model training corpus $X_{critic}$.

It is noteworthy that, to demonstrate the generality of our approach, we train a \textbf{unified} model on a mixture of all datasets, enabling it to perform reasoning across diverse multi-hop QA datasets instead of dataset-specific training.

\subsection{Retrieval}
\label{a1:retrieval}
Following prior work~\citep{ircot}, we construct a retrieval corpus using supporting facts and irrelevant documents from the training instances, and conduct retrieval within this scope.

By default, we employ BM25~\citep{bm25} provided by Elastic Search\footnote{\url{https://www.elastic.co/}} for sparse retrieval in our experiments. 
Additionally, we incorporate dense retrieval and hybrid retrieval in our analysis.
For dense retrieval, we use Contriever~\citep{contriever}\footnote{\url{https://huggingface.co/facebook/contriever-msmarco}}. 
For hybrid retrieval, we combine the results from both sparse and dense retrieval by taking their union.
Since our method independently processes each retrieved document and selects relevant ones based on self-evaluation, it inherently supports hybrid retrieval, thereby delivering better retrieval performance.

\subsection{Baselines}
\label{baselines}
\paragraph{Chain-of-Thought}~\citep{fewshot-cot} generates step-by-step reasoning before the final answer. We use 20-shot demonstrations.

\paragraph{RAG}~\citep{retrieval-then-read-1} retrieves k documents and conducts chain-of-thought reasoning conditional on the retrieval contents.
We use BM25 to retrieve 5 documents, and 3-shot demonstrations.

\paragraph{IRCoT}~\citep{ircot} interleaves retrieval and reasoning until the model arrives at the final answer. 
We set the number of documents per iteration to 4, with a maximum of 4 iterations.

\paragraph{Self-Ask}~\citep{selfask} iteratively breaks down complex questions and answers the resulting sub-questions one by one. We use 4-shot prompts.

\paragraph{ProbTree}~\citep{ptot} decomposes the question into a tree and uses bottom-up aggregation reasoning based on log-prob to get the final answer.

\paragraph{BeamAggR}~\citep{beamaggr} parses the complex question into a structured decomposition tree. It then performs bottom-up beam aggregation reasoning to select the promising reasoning trajectory.

\paragraph{Self-RAG}~\citep{selfrag} enables dynamic retrieval and generation through self-reflective tokens. We set the max retrieval documents per round to 5, and the max reasoning depth to 7.

\paragraph{RAG-Star}~\citep{rag-star} uses Monte Carlo Tree Search to guide LLM-based tree-like reasoning and employs retrieval-augmented verification to consolidate internal and external knowledge.

\paragraph{GenGround}~\citep{generate-then-ground} alternates between generating single-hop question-answer pairs using LLMs and grounding them in retrieved documents to correct potential knowledge errors.

\paragraph{Auto-RAG}~\citep{auto-rag} leverages LLMs' reasoning and decision-making capabilities to plan retrievals, query rewriting, and gather external retrieval  knowledge through multi-turn dialogues.

\paragraph{DR-Distillation}~\citep{dr-distillation} distills two small models (Decomposer and Responser) to interactively solve knowledge-intensive multi-hop questions by breaking them into simpler sub-questions and conducting retrieval-augmented reasoning.

\subsection{Task Definition}
\label{task-definition}
Given a multi-hop QA pair $(Q, ~a)$ that requires multiple reasoning steps $r_i$ to solve, each reasoning step $r_i$ is based on a specific piece of supporting evidence $d_i$.
We use the LLM to iteratively decompose the question $q_i$, employ an external retriever $\mathrm{Retr}$ to retrieve relevant documents based on the sub-question, and conduct retrieval-augmented reasoning conditional on the retrieved documents.
Once the reasoning is finished, the answer in the last iteration is regarded as the final answer $a^*$.
\begin{align}
    Q \to r_1 \to r_2 \cdots \to a^*
\end{align}

\begin{table}[h]
\centering
\setlength\tabcolsep{6pt}
\resizebox{0.95\linewidth}{!}{%
\begin{tabular}{lc}
\toprule
\bf Hyperparameters &\bf Values\\

\midrule
\multicolumn{2}{l}{\textbf{\textit{Data Synthesis Hyperparameters}}} \\
Temperature & 1.15 \\
N & 3 \\
\# Raw Generator Corpus & 49000 \\
\# Raw Critic Corpus & 6000 \\

\midrule
\multicolumn{2}{l}{\textbf{\textit{Training Hyperparameters}}} \\
Epochs & 2 \\
Learning Rate & 5e-5 \\
Batch Size & 128 \\
Warmup Ratio & 0.03 \\
LR Decay & Cosine Annealing \\
LoRA Rank & 128 \\
LoRA Alpha & 256 \\
LoRA Targets & All Linear Layers \\

\midrule
\multicolumn{2}{l}{\textbf{\textit{Inference Hyperparameters}}} \\
Temperature & 1.15 \\
Repetition Penalty & 1.2 \\
\# Docs & 5 or 6 (hybrid retrieval) \\
\# Decomposition Branches & 2 \\
\# Candidates & 2 \\
\# Retrieval Reward & 1.0 / 0.5 / -1.0 \\
\# Reasoning Reward & 0.5 / 0.25 / -0.5 \\

\bottomrule
\end{tabular}
}
\caption{Hyperparameters in experiments.}
\label{tab:hyperparameters}
\end{table}

\subsection{Implementation Details}
\label{a1:implementation-details}
The training and inference are conducted on 4 and 1 NVIDIA Tesla A100-80G GPUs, respectively.
The training process is initiated from the base model, and a chat template structured as '[INST] \{Instruction\} [/INST] \{Output\}' is manually added.
During the training process, we exclusively train the Output while masking the retrieved document portions to prevent them from being included in the optimization.
We use vLLM\footnote{\url{https://docs.vllm.ai/en/latest/}} in the inference.
Hyperparameters  are provided in Table~\ref{tab:hyperparameters}.
\clearpage
\section{Supplemental Materials on Analysis}
\label{appendix:a2}

\begin{table}[h]
\centering
\setlength\tabcolsep{4pt}
\resizebox{\linewidth}{!}{%
\begin{tabular}{lccc}
\toprule
\textbf{Methods} & \textbf{2WikiMQA} & \textbf{HotpotQA} & \textbf{MuSiQue} \\
\midrule
Self-Ask & 2.72 & 2.28 & 2.48 \\
IRCoT & 3.78 & 2.86 & 2.44 \\
Self-RAG & 2.05 & 2.12 & 2.13 \\
Auto-RAG & 2.67 & 2.39 & 3.74 \\
DR-Distillation & 3.33 & 2.85 & 3.07 \\
\ours{} & \textbf{2.40} & \textbf{2.01} & \textbf{2.37} \\
\bottomrule
\end{tabular}%
}
\caption{
The average number of iterations of iterative reasoning methods to solve a multi-hop question.
Our method requires fewer iterations to solve a multi-hop question, resulting in higher reasoning efficiency.
}
\label{tab:analysis_efficiency}
\end{table}
\begin{table}[h]
\centering
\setlength\tabcolsep{4pt}
\resizebox{1.0\linewidth}{!}{%
\begin{tabular}{lcccc}
\toprule
\textbf{Question Type} & \textbf{Bdg.} & \textbf{Infer.} & \textbf{Cmp.} & \textbf{Bdg-Cmp.} \\
\midrule
Latency(s) & 2.41 & 2.56 & 2.91 & 5.56 \\
Prompt & 2989.4 & 2775.0 & 1899.1 & 4653.9 \\
Completion & 171.5 & 202.4 & 209.3 & 375.2 \\
\bottomrule
\end{tabular}
}
\caption{
Per-sample inference latency and token consumption of greedy inference on 2WikiMQA dataset.
}
\label{tab:analysis_efficiency_2wikimqa}
\end{table}

\begin{table}[h]
\centering
\setlength\tabcolsep{4pt}
\resizebox{0.7\linewidth}{!}{%
\begin{tabular}{lcc}
\toprule
\textbf{Question Type} & \textbf{Bridge} & \textbf{Comparison} \\
\midrule
Latency(s) & 3.71 & 2.81 \\
Prompt & 2156.6 & 1786.6 \\
Completion & 182.7 & 220.3 \\
\bottomrule
\end{tabular}
}
\caption{
Per-sample inference latency and token consumption of greedy inference on HotpotQA dataset.
}
\label{tab:analysis_efficiency_hotpotqa}
\end{table}

\begin{table}[h]
\centering
\setlength\tabcolsep{4pt}
\resizebox{0.8\linewidth}{!}{%
\begin{tabular}{lccc}
\toprule
\textbf{Question Type} & \textbf{2hop} & \textbf{3hop} & \textbf{4hop} \\
\midrule
Latency(s) & 2.43 & 3.41 & 4.67 \\
Prompt & 2388.9 & 4162.6 & 4294.4 \\
Completion & 182.1 & 258.2 & 333.8 \\
\bottomrule
\end{tabular}
}
\caption{
Per-sample inference latency and token consumption of greedy inference on MuSiQue dataset.
}
\label{tab:analysis_efficiency_musique}
\end{table}

\subsection{Reasoning Efficiency between Separated Critic and Generative Critic}
\label{a2:reward-modeling}
As mentioned in \S\ref{analysis:reward_modeling}, GenCritic achieves a higher throughput compared to Separated Critic.
We will describe how to calculate the throughput in this section.
We use the same computing platform, equipped with one A100-SXM4-80GB GPU, a Xeon(R) 6348 CPU with 28 cores and 56 threads, and 480GB of memory.
We sample 500 instances from each of the three datasets, totaling 1,500 instances.
We conduct parallel inference using the API server provided by vLLM, employing reward-guided greedy inference during benchmarking.
The throughput is calculated based on the time consumed for different methods.

\subsection{Iterative Reasoning Efficiency}
\label{a2:iterative-reasoning-efficiency}

We discuss the iterative reasoning efficiency in \S\ref{analysis:efficiency}.
Specifically, we consider a complete round of retrieval-generation as one reasoning iteration. 
It is noteworthy that multiple retrieval calls can occur within a single reasoning iteration.
The specific number of reasoning iterations corresponding to each method can be found in Table~\ref{tab:analysis_efficiency}.
Additionally, we present the inference latency of individual samples with greedy inference, along with the average prompt and completion token costs, as shown in Table~\ref{tab:analysis_efficiency_2wikimqa}, Table~\ref{tab:analysis_efficiency_hotpotqa},
and Table~\ref{tab:analysis_efficiency_musique}.
\begin{table*}[h]
\centering
\begin{tabularx}{\linewidth}{cX}
\toprule
Notation  & Description \\
\midrule
\multicolumn{2}{l}{\textit{\textbf{(a) Modules}}} \\
$LLM$ & The advanced large language model to synthesis iterative reasoning rationales and process supervision. \\
$R$ & The iterative reasoner model. \\
$C$ & The separated critic model. \\
$R_{sc}$ & The iterative reasoner model with self-critique capabilities.  \\
$\mathrm{Retr}$  & External retriever (sparse, dense, hybrid) \\

\midrule
\multicolumn{2}{l}{\textit{\textbf{(b) Training Corpus}}} \\
$X$ &  Raw training corpus for generator. \\
$X_{critic}$ &  Raw training corpus for critic, which is disjoint with $X$. \\

\midrule
\multicolumn{2}{l}{\textbf{(c) Symbols}} \\
$q^{(r)}_i$  &  Multi-hop question after reduction. \\
$q_{i}$ & Decomposed atomic sub-question. \\
$d_i$ & The retrieved document. \\
$x$ & Reasoning trajectory. \\
$s_{i}$ & Sub-question reasoning. \\
$r_{i}^{\mathrm{retr}}$ & Reward for retrieval relevance. \\
$r_{i}^{\mathrm{reas}}$ & Reward for reasoning utility. \\
$r_{i}^{c}$ & Cumulative process reward. \\
$r_i^{o}$ & Outcome reward. \\

\bottomrule
\end{tabularx}
\caption{
The formal notations mentioned in Section~\ref{sec:methods}.
}
\label{tab:notations}
\end{table*}

\clearpage
\section{Prompts}
\label{appendix:a3}

We provide the manually annotated prompts for data synthesis.
For iterative reasoning rationales, we use 1-shot prompts for each question type (Figure~\ref{demo:generator}).
For critic on retrieval and reasoning, we use 3-shot prompts (Figure~\ref{demo:critic-retrieval}, \ref{demo:critic-reasoning}).
For critic on outcome reward, we use zero-shot prompts (Figure~\ref{demo:critic_outcome}).

Additionally, we provide an example of iterative reasoning with self-critique, shown in Figure~\ref{demo:reasoning_examples}.

\begin{algorithm*}[t]
\begin{algorithmic}[1]
\Require Initial question $q_0$, retrieval model $\mathrm{Retr}$, probabilistic language model $p$, candidate size $k$
\Ensure Final answer $a^*$

\Statex \textbf{Initialize:}
\State $\vx_0 \gets [q_0]$ \Comment{Initial reasoning prefix}
\State $\vr^{\mathrm{c}}_0 \gets 0$ \Comment{Initial c reward}
\State $\vx_0 \gets \{(\vx_0, \vr^{\mathrm{c}}_0)\}$ \Comment{Set of candidate trajectories}

\Statex \textbf{Main Loop:}
\For{$t = 0$ to $T-1$} \Comment{$T$ is the maximum number of iterations}
    \State $\vx_{t+1} \gets \emptyset$ \Comment{Initialize set for next iteration}
    
    \Statex \textbf{Branch Exploration:}
    \For{$(\vx_t, \vr^{\mathrm{c}}_t)$ in $\vx_t$}
        
        \State \textbf{Decomposition:}
        \State $q_{t+1} \sim p(y \mid q_t^{(r)}, \vx_t)$ \Comment{Sample sub-question}
        
        \State \textbf{Retrieval:}R
        \State $d_i \gets \mathrm{Retr}(q_{t+1})$ \Comment{Retrieve relevant documents}
        \State $r_{t+1}^{\mathrm{retr}} \gets p(y \mid q_{t+1}, d_i)$ \Comment{Compute retrieval reward}
        
        \State \textbf{Reasoning:}
        \State $s_{t+1} \sim p(y \mid q_{t+1}, d_i, \vx_t)$ \Comment{Sample reasoning step}
        \State $r_{t+1}^{\mathrm{reas}} \gets p(y \mid q_{t+1}, d_i, s_{t+1})$ \Comment{Compute reasoning reward}
        
        \State \textbf{Update Reasoning Prefix:}
        \State $\vx_{t+1} \gets \vx_t \oplus [q_{t+1}, d_i, r_{t+1}^{\mathrm{retr}}, s_{t+1}, r_{t+1}^{\mathrm{reas}}]$ \Comment{Concatenate current status}
        
        \State \textbf{Update Cumulative Reward:}
        \State $\vr^{\mathrm{c}}_{t+1} \gets \vr^{\mathrm{c}}_t + r_{t+1}^{\mathrm{retr}} + r_{t+1}^{\mathrm{reas}}$

        \State \textbf{Question Reduction:}
        \State $q_t^{(r)} \gets \text{ReduceQuestion}(\vx_t)$ \Comment{Reduce remaining question}
        
        \State $\vx_{t+1} \gets \vx_{t+1} \cup \{(\vx_{t+1}, \vr^{\mathrm{c}}_{t+1})\}$ \Comment{Add new candidate}
    \EndFor
    
    \Statex \textbf{Candidate Selection:}
    \State $\vx_{t+1} \gets \text{Top-k}(\vx_{t+1}, \vr^{\mathrm{c}}_{t+1})$ \Comment{Select top-k candidates}
\EndFor

\Statex \textbf{Final Answer Selection:}
\State $a^* \gets \arg\max_{a} \left( \sum_{i=1}^n (r_i^{\mathrm{retr}} + r_i^{\mathrm{reas}}) \right)$ \Comment{Select answer with highest c reward}
\State \Return $a^*$
\end{algorithmic}
\caption{Self-Critique Guided Iterative Reasoning}
\label{alg:sc-reasoner-inference}
\end{algorithm*}
\begin{algorithm*}[t]
\begin{algorithmic}[1]
\Require Generator corpus $X = \{x_i\}$, where $x_i = (q_i, a_i, \{d_1, \dots, d_n\})$
\Require Non-overlapping critic corpus $X_{critic}$
\Require Large Language Model (LLM)
\Ensure Self-Critique Reasoner $R_{sc}$

\Statex \textbf{Phase 1: Train Iterative Reasoner $R$}
\State $X_{\text{synthetic}} \gets \emptyset$ \Comment{Initialize synthetic reasoning corpus}
\For{each instance $x_i \in X$}
    \State \textbf{Step 1.1: Synthesize Reasoning Rationales}
    \State $\text{decomp}_i, \text{reasoning}_i \gets \text{LLM}(q_i, \{d_1, \dots, d_n\})$ \Comment{Few-shot prompting for decomposition and reasoning}
    \State $\text{trajectory}_i \gets \text{Organize}(\text{decomp}_i, \text{reasoning}_i)$ \Comment{Organize rationales}
    \State $X_{\text{synthetic}} \gets X_{\text{synthetic}} \cup \{\text{trajectory}_i\}$
\EndFor

\State \textbf{Step 1.2: Train Iterative Reasoner}
\State $R \gets \text{TrainModel}(X_{\text{synthetic}})$ \Comment{Train $R$ on synthetic corpus}

\\
\Statex \textbf{Phase 2: Train Critic Model $C$}
\State $X_{\text{critic-data}} \gets \emptyset$ \Comment{Initialize critic training data}
\For{each instance $x_i \in X_{critic}$}
    \State \textbf{Step 2.1: Generate Reasoning Trajectories}
    \State $\text{trajectory}_i \gets R(q_i, \{d_1, \dots, d_n\})$ \Comment{Use $R$ to generate reasoning trajectories}
    \State \textbf{Step 2.2: Evaluate Rewards with LLM}
    \State $r^{\mathrm{retr}}_i, r^{\mathrm{reas}}_i, r^{\mathrm{overall}}_i \gets \text{LLM}(\text{trajectory}_i)$ \Comment{Evaluate retrieval, reasoning, and overall quality}
    \State $X_{\text{critic-data}} \gets X_{\text{critic-data}} \cup \{(\text{trajectory}_i, r^{\mathrm{retr}}_i, r^{\mathrm{reas}}_i, r^{\mathrm{overall}}_i)\}$
\EndFor
\State \textbf{Step 2.3: Train Critic Model}
\State $C \gets \text{TrainModel}(X_{\text{critic-data}})$ \Comment{Train $C$ on critic data}

\\
\Statex \textbf{Phase 3: Train Self-Critique Reasoner $R_{sc}$}
\State $X_{\text{sc-data}} \gets \emptyset$ \Comment{Initialize self-critique training data}
\For{each instance $x_i \in X$}
    \State \textbf{Step 3.1: Annotate Reasoning Trajectories with Critic}
    \State $\text{trajectory}_i \gets R(q_i, \{d_1, \dots, d_n\})$ \Comment{Generate reasoning trajectories with $R$}
    \State $r^{\mathrm{retr}}_i, r^{\mathrm{reas}}_i, r^{\mathrm{overall}}_i \gets C(\text{trajectory}_i)$ \Comment{Annotate with critic $C$}
    \State $\text{trajectory}_i^{\text{annotated}} \gets \text{AddSpecialTokens}(\text{trajectory}_i, r^{\mathrm{retr}}_i, r^{\mathrm{reas}}_i, r^{\mathrm{overall}}_i)$
    \State $X_{\text{sc-data}} \gets X_{\text{sc-data}} \cup \{\text{trajectory}_i^{\text{annotated}}\}$
\EndFor
\State \textbf{Step 3.2: Train Self-Critique Reasoner}
\State $R_{sc} \gets \text{TrainModel}(X_{\text{sc-data}})$ \Comment{Train $R_{sc}$ on annotated corpus}

\\
\Statex \textbf{Self-Improvement Pipeline}
\While{not converged}
    \State \textbf{Step 4.1: Generate Synthetic Rationales}
    \State $X_{\text{new}} \gets R_{sc}(\text{unlabeled questions})$ \Comment{Generate rationales for unlabeled questions}
    \State \textbf{Step 4.2: Evaluate Rationale Quality}
    \State $X_{\text{high-quality}} \gets \{x \in X_{\text{new}} \mid C(x) > \text{threshold}\}$ \Comment{Filter high-quality rationales}
    \State \textbf{Step 4.3: Retrain $R_{sc}$}
    \State $R_{sc} \gets \text{TrainModel}(R_{sc}, X_{\text{high-quality}})$ \Comment{Retrain with high-quality data}
\EndWhile

\State \Return $R_{sc}$
\end{algorithmic}
\caption{Training SC-Reasoner}
\label{alg:sc-reasoner-training}
\end{algorithm*}
\definecolor{mygreen}{HTML}{229954}
\definecolor{mybrown}{HTML}{BA4A00}

\begin{table*}[t]
\centering
\resizebox{\textwidth}{!}{
\begin{tabularx}{\linewidth}{cX}
\toprule
\multicolumn{2}{c}{\textbf{Iterative Reasoning Examples of Self-Critique Reasoner}} \\ 

\midrule
\multirow{2}{*}{\textbf{Q}} & How many counties are there in the largest state in the region where the film The Family Stone was set? (Answer: sixteen)\\
\midrule
\multirow{26}{*}{\textbf{A}} & [Non-Atomic Question] \newline <sub-question> In which region is the film "The Family Stone" set?\newline <paragraph>The Family Stone ...</paragraph>\newline Retrieval Quality: \textcolor{mybrown}{[Partially Relevant]}\newline
Reasoning: From Document, we know that "The Family Stone" is set in a small New England town. Therefore, the region is New England.\newline</sub-question> Reasoning Quality: \textcolor{mygreen}{[Fully supported]}\newline [Remaining Question] How many counties are there in the largest state in New England? \newline
\newline [Non-Atomic Question] \newline <sub-question> Which state is the largest in New England?\newline <paragraph>New England ...</paragraph>\newline Retrieval Quality: \textcolor{mybrown}{[Partially Relevant]}\newline Reasoning: From Document, we know that Maine is the largest state in New England, constituting nearly one-half of the total area of New England.\newline </sub-question> Reasoning Quality: \textcolor{mygreen}{[Fully supported]} \newline [Remaining Question] How many counties are there in Maine? \newline
\newline[Atomic Question]\newline<paragraph>List of counties in Maine ...</paragraph>\newline Retrieval Quality: \textcolor{mygreen}{[Relevant]}\newline Reasoning: From Document, we know that there are sixteen counties in Maine.\newline Reasoning Quality: \textcolor{mygreen}{[Fully supported]} \newline
[Final Answer] @@\textbf{sixteen}@@\newline
Overall Reasoning Quality: \textcolor{mygreen}{[4]}\\
\bottomrule
\end{tabularx}
}
\caption{
A reasoning example of self-critique reasoner.
The data instance is from MuSiQue dataset.
}
\label{demo:reasoning_examples}
\end{table*}

\begin{figure*}[h]
    \centering
    \promptfigure{Prompts for \textbf{Iterative Reasoning Rationales}}{
Please answer the following multi-hop questions with explanation based on the given documents.  Wrap the final answer with @@ @@. Here is an example: \newline

\#\#\# Document \#1: Cuidado con las imitaciones \newline
Cuidado con las imitaciones is a 1948 Argentine film directed by Luis Bayón Herrera. \newline

\#\#\# Document \#2: Luis Bayón Herrera \newline
Luis Bayón Herrera (23 September 1889 – 30 March 1956) was a Spanish film director and screenwriter who worked in Argentine film of the 1940s and 1950s. He was \"one of the most important directors of the golden age of Argentine cinema\". Herrera was born in Bilbao, Spain. He directed some 40 different Argentine feature films and in the last few years of his career in the early 1950s he worked on Cuban production with films such as \"A La Habana me voy\" in 1951. He died in Buenos Aires in 1956, aged 66.", \newline

\#\#\# Question: What nationality is the director of film Cuidado Con Las Imitaciones? \newline

To answer this multi-hop question, we need to solve the following sub-questions: \newline

\#\#\# Decompose the multi-hop question into sub-questions \newline
1. Who is the director of the film Cuidado Con Las Imitaciones? \newline
2. What is the nationality of that director? \newline

Next, let's solve the sub-questions one by one. \newline

\#\#\# Sub-question: Who is the director of the film Cuidado Con Las Imitaciones? \newline
From Document \#1, we know that "Cuidado Con Las Imitaciones" is a Spanish film, and its director is José María Forqué. \newline

\#\#\# Remaining Question: What is the nationality of José María Forqué? \newline
From Document \#2, we know that José María Forqué was a Spanish director, which means his nationality is Spanish. \newline

\#\#\# Final Answer \newline
@@Spanish@@ \newline

\{SUPPORTING DOCUMENTS\} \newline

\#\#\# Question: \{MULTI-HOP QUESTION\}
}
    \caption{The fewshot prompts for iterative reasoning rationales synthesis.}
    \label{demo:generator}
\end{figure*}
\begin{figure*}[h]
    \centering
    \promptfigure{Prompts for \textbf{Retrieval Relevance Evaluation}}{
You will be given a question and an evidence document. Your task is to evaluate whether the evidence is relevant to the question and provide a clear explanation for your decision. Use the following scale to rate the relevance of the evidence:\newline
[Relevant]: The evidence directly addresses or supports the question.\newline
[Partially Relevant]: The evidence is somewhat related to the question but does not fully address it. For example, it may mention the core entity or topic of the question.\newline
[Irrelevant]: The evidence does not relate to the question in any meaningful way.\newline

\#\#\# Question: In what country was the first Pan-African conference held?\newline

\#\#\# Evidence: First Pan-African Conference\newline
The First Pan-African Conference was held in London from 23 to 25 July 1900 (just prior to the Paris Exhibition of 1900 in order to allow tourists of African descent to attend both events ). Organized primarily by the Trinidadian barrister Henry Sylvester Williams, it took place in Westminster Town Hall (now Caxton Hall) and was attended by 37 delegates and about 10 other participants and observers from Africa, the West Indies, the US and the UK, including Samuel Coleridge Taylor (the youngest delegate), John Alcindor, Dadabhai Naoroji, John Archer, Henry Francis Downing, and W.E.B. Du Bois, with Bishop Alexander Walters of the AME Zion Church taking the chair.\newline

Rating: [Relevant]\newline
Explanation: The evidence directly answers the question by stating that the first Pan-African conference was held in London, which is in the United Kingdom. The details provided, such as the dates and location of the conference, further confirm the relevance of the evidence to the question.\newline

\#\#\# Question: What is the administrative territorial entity for Malta Township?\newline

\#\#\# Evidence: Minsk Region\newline
Minsk Region or Minsk Voblasć or Minsk Oblast (, "Minskaja vobłasć" ; , "Minskaja oblastj") is one of the regions of Belarus. Its administrative center is Minsk, although it is a separate administrative territorial entity of Belarus. As of 2011, the region's population is 1,411,500.\newline

Rating: [Irrelevant]\newline
Explanation: The evidence provided discusses the Minsk Region, which is a region in Belarus. The question, however, asks about the administrative territorial entity for Malta Township. Since the evidence pertains to Belarus and not Malta, it is not relevant to the question.\newline

\#\#\# Question: Did Bob Rosa produce any work for Patrick Love?\newline

\#\#\# Evidence: Bob Rosa\newline
Bob Rosa is an American record producer that has worked with artists such as Madonna, Janet Jackson, Mariah Carey and Espen Lind. One if his first recording projects was working on the track "Planet Rock" by Afrika Bambaataa \& the Soulsonic Force in 1982.\newline

Rating: [Partially Relevant]\newline
Explanation: The evidence mentions Bob Rosa's work with various artists, including Madonna, Janet Jackson, Mariah Carey, and Espen Lind. However, it does not specifically mention Patrick Love or any work Bob Rosa may have done for him. Therefore, the evidence is partially relevant to the question.\newline

\#\#\# Question: \{\}\newline

\#\#\# Evidence: \{\}\newline
}
    \caption{The fewshot prompts for retrieval quality critic.}
    \label{demo:critic-retrieval}
\end{figure*}
\begin{figure*}[h]
    \centering
    \promptfigure{Prompts for \textbf{Reasoning Utility Evaluation}}{
You will be provided with a question, an evidence document, along with a response.
Your job is to determine whether the response is supported by the evidence, and provide explanation for your decision.
Use the following scale to rate the response:\newline
[Fully supported] Most of the information in the response is supported by the evidence.\newline
[Partially supported] Some of the information in the response is supported by the evidence, but there are some parts that are speculative.\newline
[Not supported] The response is not supported by the evidence / The response does not provide useful reasoning.\newline

\#\#\# Question: In which country is the Financial Conduct Authority located?\newline

\#\#\# Evidence: Martin Wheatley\newline
Martin Wheatley is a British financier, formerly managing director of the Consumer and Markets Business Unit of the Financial Services Authority in the UK, and is the former CEO of the Financial Conduct.\newline

\#\#\# Response: From Document, we know that Martin Wheatley is a British financier who formerly managed the Consumer and Markets Business Unit of the Financial Services Authority in the UK. This indicates that the Financial Conduct Authority is located in the United Kingdom.\newline

Rating: [Fully supported]\newline
Explanation: The response is fully supported by the evidence. The evidence clearly states that Martin Wheatley is a British financier who formerly managed the Consumer and Markets Business Unit of the Financial Services Authority in the UK and was the former CEO of the Financial Conduct Authority. The response accurately extracts this information and concludes that the Financial Conduct Authority is located in the United Kingdom, which is directly supported by the evidence provided.\newline

\#\#\# Question: What is the birthplace of Jane Siberry?\newline

\#\#\# Evidence: Gavin Bradley \newline
Gavin Bradley is an award-winning Canadian singer-songwriter and producer based in Toronto who has worked with artists like Nelly Furtado, Tori Amos and Jane Siberry. Fusing acoustic and electronic elements, his work is identifiable for its signature "warm" piano sound and live strings mixed with filtered synthesizers and other electronic manipulations . Besides production, Bradley is a solo recording artist. His debut album 'Deep Freeze' was released on UMI Records in 2006.\newline

\#\#\# Response: The response is partially supported because the evidence mentions Gavin Bradley's Toronto connection but does not directly state Jane Siberry's birthplace. The inference about her possible connection to Toronto is speculative.\newline

Rating: [Partially supported]\newline
Explanation: The response correctly identifies that the evidence mentions Gavin Bradley's connection to Toronto but does not directly state Jane Siberry's birthplace. The inference that Jane Siberry might be connected to Toronto based on her collaboration with Gavin Bradley is speculative, as the evidence does not provide explicit information about her birthplace. Therefore, the response is partially supported.\newline

\#\#\# Question: \{\}\newline

\#\#\# Evidence: \{\}\newline

\#\#\# Response: \{\}\newline
}
    \caption{The fewshot prompts for reasoning utility critic.}
    \label{demo:critic-reasoning}
\end{figure*}
\begin{figure*}[h]
    \centering
    \promptfigure{Prompts for \textbf{Outcome Reward Evaluaion}}{
You will be provided with a question, along with a reasoning trajectory.\newline
Your job is to determine whether the reasoning trajectory is useful for answering the question, and provide explanation for your decision.\newline
Use the following scale to rate the reasoning:\newline
[5]: The reasoning process is clear, logically structured, and well-supported by the evidence.\newline
[4]: The reasoning process is mostly clear, partially logically structured, and supported by evidence, but may contains minor logical flaws.\newline
[3]: The reasoning is somewhat unclear with noticeable flaws and uses limited or weak evidence.\newline
[2]: The reasoning is flawed, lacking supporting evidence, which results in an incorrect conclusion.\newline
[1]: The reasoning fails to provide a definitive answer.\newline

\#\#\# Question: \{\}\newline

\#\#\# Reasoning: \{\}\newline

Rating: \newline
Explanation:
}
    \caption{The prompts for reasoning trajectories outcome reward evaluation.}
    \label{demo:critic_outcome}
\end{figure*}

\end{document}